\documentclass[sigconf]{acmart}

\renewcommand\footnotetextcopyrightpermission[1]{} % removes footnote with conference information in first column
\AtBeginDocument{%
  \providecommand\BibTeX{{%
    \normalfont B\kern-0.5em{\scshape i\kern-0.25em b}\kern-0.8em\TeX}}}

\setcopyright{acmcopyright}
\copyrightyear{2018}
\acmYear{2018}
\acmDOI{XXXXXXX.XXXXXXX}

\acmConference[MM '22]{Make sure to enter the correct
  conference title from your rights confirmation emai}{October 10-14,
  2022}{ Lisbon, Portugal}
\acmPrice{15.00}
\acmISBN{978-1-4503-XXXX-X/18/06}

\usepackage{url}            % simple URL typesetting
\usepackage{booktabs}       % professional-quality tables
\usepackage{amsfonts}       % blackboard math symbols
\usepackage{nicefrac}       % compact symbols for 1/2, etc.
\usepackage{microtype}      % microtypography
\usepackage{xcolor}         % colors

\usepackage[noend]{algpseudocode}
\usepackage{booktabs}
\usepackage{multirow}
\usepackage{algorithmicx,algorithm}
\usepackage{pifont}
\usepackage{array}
\usepackage{enumitem}
\usepackage{makecell}
\usepackage{graphicx}
\usepackage{soul}
\usepackage{wrapfig}
\usepackage{color}

\newlength{\bibitemsep}
\newlength{\bibparskip}
\let\oldthebibliography\thebibliography
\renewcommand\thebibliography[1]{%
  \oldthebibliography{#1}%
  \setlength{\parskip}{\bibitemsep}%
  \setlength{\itemsep}{\bibparskip}%
}

\newcommand{\PreserveBackslash}[1]{\let\temp=\\#1\let\\=\temp}
\newcolumntype{C}[1]{>{\PreserveBackslash\centering}p{#1}}
\newcolumntype{R}[1]{>{\PreserveBackslash\raggedleft}p{#1}}
\newcolumntype{L}[1]{>{\PreserveBackslash\raggedright}p{#1}}

\begin{document}

\title{Instant Response Few-shot Object Detector via Meta-Strategy and Explicit Localization Inference}


\author{Junying~Huang,
        Fan~Chen,
        Sibo~Huang
        and~Dongyu~Zhan}

% \affiliation{
%     \institution{Paper under double-blind review}
% \renewcommand{\shortauthors}{Anonymous Author, et al.}


\begin{abstract}
Aiming at recognizing and localizing the object of novel categories by a few reference samples, few-shot object detection (FSOD) is a quite challenging task. Previous works often depend on the fine-tuning process to transfer their model to the novel category and rarely consider the defect of fine-tuning, resulting in many application drawbacks. For example, these methods are far from satisfying in the episode-changeable scenarios due to excessive fine-tuning times, and their performance on low-quality (e.g., low-shot and class-incomplete) support sets degrades severely. To this end, this paper proposes an instant response few-shot object detector (IR-FSOD) that can accurately and directly detect the objects of novel categories without the fine-tuning process. To accomplish the objective, we carefully analyze the defects of individual modules in the Faster R-CNN framework under the FSOD setting and then extend it to IR-FSOD by improving these defects. Specifically, we first propose two simple but effective meta-strategies for the box classifier and RPN module to enable the object detection of novel categories with instant response. Then, we introduce two explicit inferences into the localization module to alleviate its over-fitting to the base categories, including explicit localization score and semi-explicit box regression. Extensive experiments show that the IR-FSOD framework not only achieves few-shot object detection with the instant response but also reaches state-of-the-art performance in precision and recall under various FSOD settings.

\end{abstract}



%%
% \ccsdesc


\maketitle

\section{Introduction}
\label{sec: introduction}

In the past decade, deep-learning-based methods have achieved remarkable success in various computer vision tasks, such as image classification \cite{resnet, imagenet} and object detection \cite{f-rcnn, yolov1}. However, these methods are fundamentally dependent on a large amount of annotated data, so their generalization ability toward the open-world tasks is limited. This triggers active research on few-shot learning, which aims to develop models that can be generalized to the unseen categories with only a few support data with annotations.

\begin{figure}
    \centering
    % \vspace{-15pt}
    \includegraphics[width=\columnwidth]{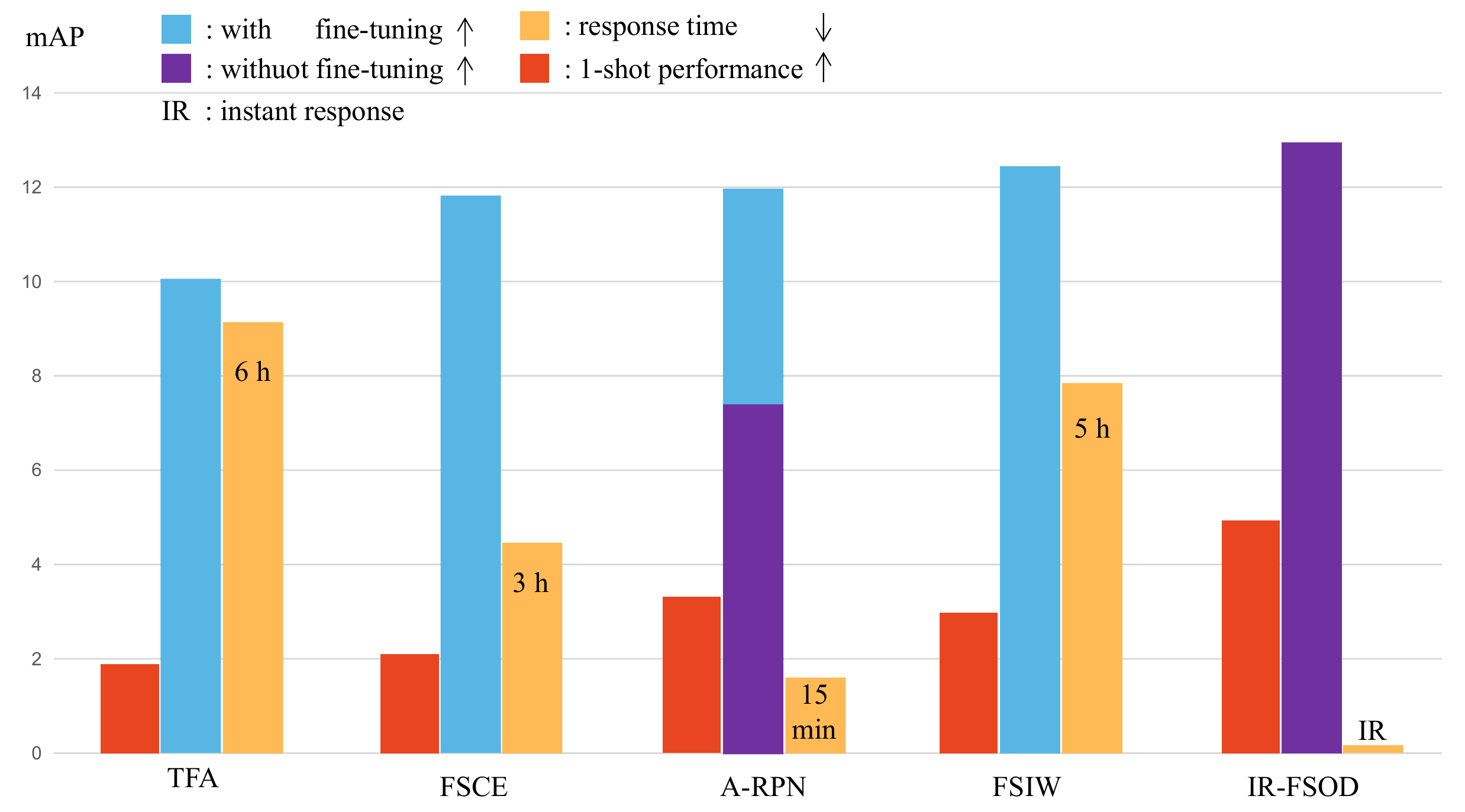}
    \vspace{-20pt}
    \caption{Comprehensive comparison between different models on MS COCO dataset, including response time for fine-tuning, performance with/without fine-tuning, and the performance under low-shot setting. TFA \cite{tfa}, FSIW \cite{fsiw}, and FSCE \cite{fsce} are invalidated before fine-tuning. A-RPN \cite{fsod} can be respond instantly the same as ours, but it performs poorly before fine-tuning. On the contrary, IR-FSOD can achieved optimal performance while supporting instant response.
    }
    \vspace{-10pt}
\label{fig: tuning_time}
\end{figure}

By leveraging meta-learning and distance metric learning, remarkable progress has been made in few-shot image classification. However, the works attempting to solve few-shot object detection (FSOD) have encountered setbacks due to the complexity of object detection tasks. Most existing methods, whether meta-learning-based or fine-tuning-based, require a cumbersome fine-tuning process on the support data. Otherwise, they will suffer from seriously degraded or even be invalidated (e.g., the methods mentioned in Figure \ref{fig: tuning_time}). 
Therefore, these methods still have some drawbacks: (a) Due to the need for independent fine-tuning processes in each episode, they are far from satisfying in the episode-changeable scenario, requiring unacceptable response times. (b) They tend to ignore the instances diverse from the support set due to over-fitting to the support instances, resulting in a low recall. (c) The effectiveness of fine-tuning depends on the quality of support sets, so they perform poorly under settings such as low-shot and class-incomplete.

To remedy these defects, this paper proposes a novel Instant Response Few-shot Object Detection (IR-FSOD) framework. ``Instant response'' (IR) refers specifically to the direct object detection for the novel category without preparatory work such as fine-tuning. The IR-FSOD framework is based on Faster R-CNN \cite{f-rcnn} and built by addressing its defects in few-shot object detection, including the improvements on the box classifier, RPN module, and the localization module in the R-CNN derived model.

First of all, the box multi-classifier can only classify the region proposal into the seen category, which directly causes many existing few-shot object detectors \cite{tfa, fsce, fsiw} to be invalidated on the novel category before fine-tuning. A-RPN \cite{fsod} and RepMet \cite{repmet} attempt to replace the multi-classifier with the comparison-classifier and distance-classifier to recognize the instance of novel categories. However, object detection is a much more complex task, which requires detecting the foreground, localization, and classification. The comparison-classifier and distance-classifier perform poorly in learning the complex feature space for object detection since they can't preserve the class-specific knowledge, so the performance of RepMet and A-RPN lags behind the fine-tuning-based methods.

\begin{figure}[t]
    \centering
    \includegraphics[width=0.8\columnwidth]{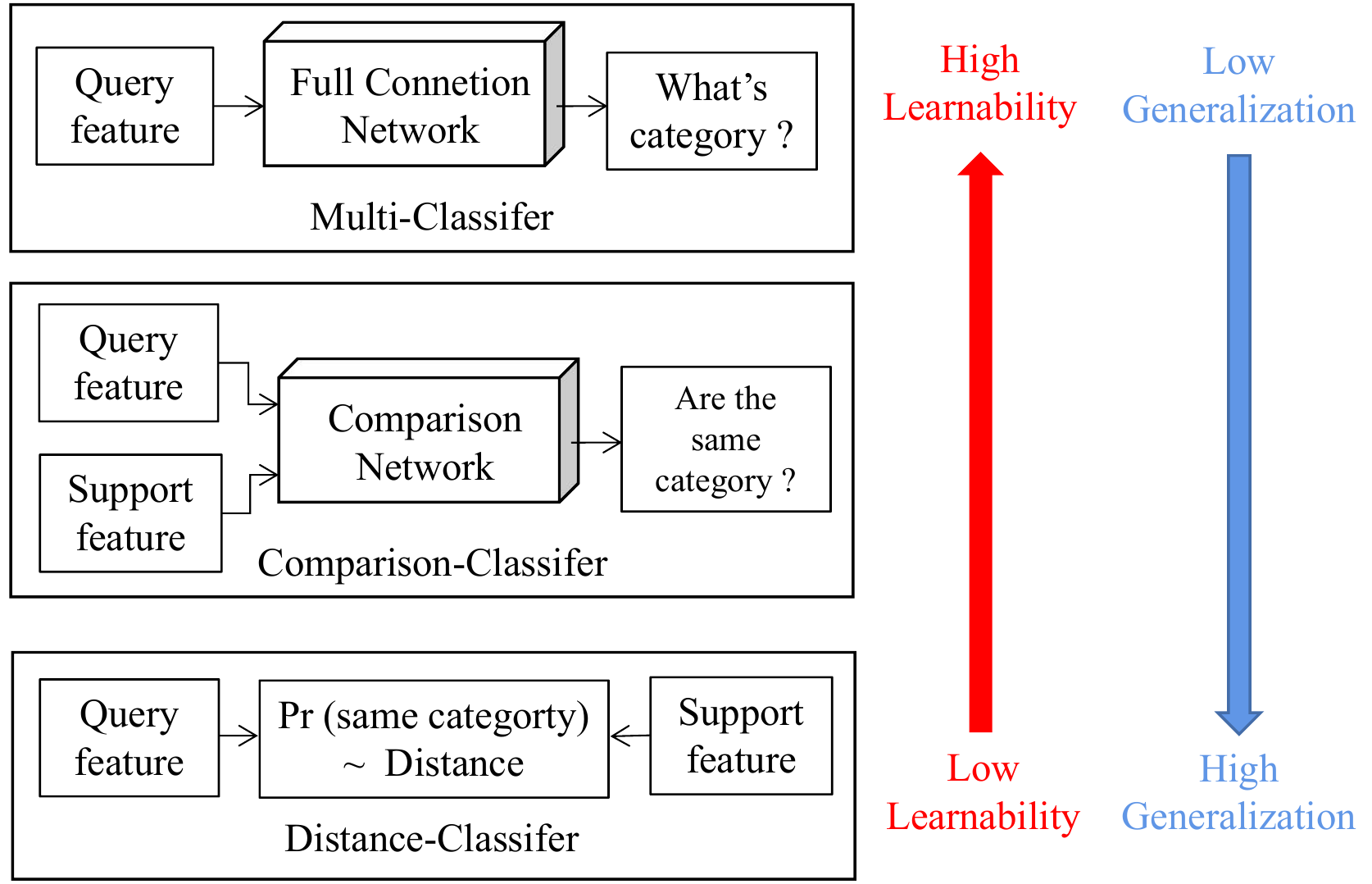}
    \vspace{-5pt}
    \caption{ Comparisons of the motivation, learnability, and generalization ability between three classifiers. 
    }
    \vspace{-15pt}
\label{fig: motivation_dyn}
\end{figure}

In Figure \ref{fig: motivation_dyn}, we compare the three classifiers mentioned above:
(a) The multi-classifier explicitly learns a hyper-plane for each base category on the feature space, which shows the best learnability but is invalidated on the novel category; (b) The comparison-classifier learns a class-agnostic binary classifier on the joint feature space. It can directly recognize the objects of novel categories but still suffers from the bias of base categories since its parameters are trained on the base category data; (c) The distance-classifier is non-parameter and performs classification according to the nearest neighbor rule. Although it can't preserve the knowledge from training, it doesn't suffer from category bias.
Based on the observation, we propose a meta-strategy called the dynamic classifier module that uses different classifiers during training and inference to build a box classifier with both generalization and learnability in the IR-FSOD framework. It can significantly improve the few-shot performance while supporting instant response.
% While getting rid of fine-tuning, it also significantly improves the few-shot performance. 

\begin{figure}[t]
    \centering
    \includegraphics[width=0.475\columnwidth]{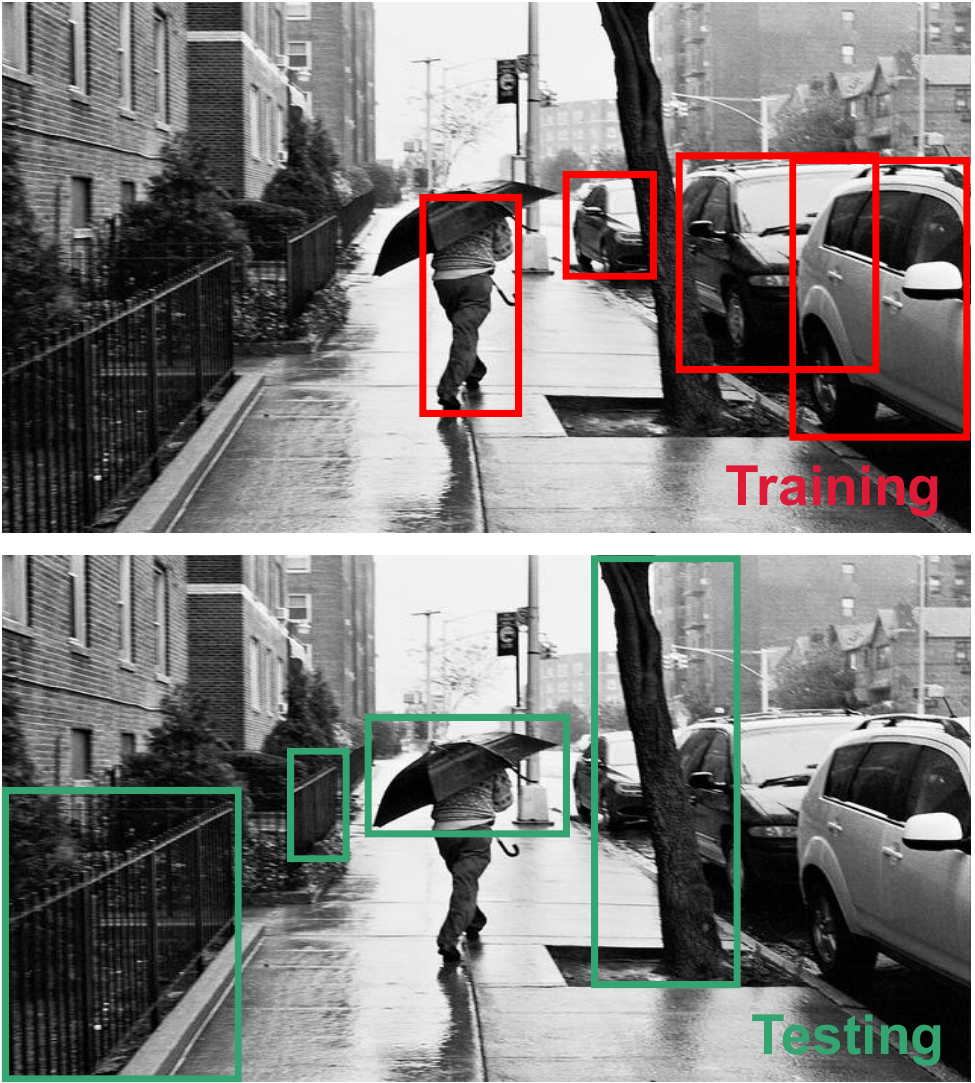}
    \vspace{-10pt}
    \caption{ The positive regions for the RPN in the training and testing phase are disjoint.
    }
    \vspace{-15pt}
\label{fig: motivation_rpn}
\end{figure}

Second, the region proposal network (RPN) suffers a fatal defect in few-shot object detection. As shown in Figure \ref{fig: motivation_rpn}, the positive regions in the training phase are disjoint with the related regions for the novel category so that these related regions are either ignored or regarded as background in the base training. A-RPN \cite{fsod} and meta-rcnn \cite{meta-rcnn} proposes to generate class-specific region proposals to avoid RPN focusing only on the base class. But the fundamental problem that only the objects belonging to base categories are treated as positive anchors in base training remains unresolved, which leads to the bottleneck of their models before fine-tuning. FSCE \cite{fsce} and R-FSOD \cite{gfsod} also point out the defects of the RPN module, but they only adjust the RPN module in the fine-tuning process, increasing the dependence on the fine-tuning process.

Different from them, we argue that the RPN module should focus on any potential foreground instances during base training instead of only the annotated instances. Therefore, we propose another meta-strategy that trains the RPN module with semi-supervised algorithms to capture the potential foreground instances. Concretely, the negative anchors during the base training are actually composed of background and the potential object instances not belonging to the base categories, so we remark them as unlabeled data and leverage them by the semi-supervised method.

% class-specific

% Third, in few-shot learning, only a few annotations for novel categories are supported. However, the localization module in the R-CNN derived model, composed of the localization score and box regression, is enabled by implicit fitting and without logic inference, so it depends on training by a large amount of annotated data.
Third, the localization module in the R-CNN derived model, composed of the localization score and box regression, is enabled by implicit fitting and lacks logic inference, which suffers from over-fitting to the distribution of the base categories. Therefore, we propose to improve the generalization ability of the localization module by introducing explicit logic inference. Concretely, we first introduce the pixel-wise contrast into the box classifier to explicitly evaluate the localization score, which can generate the confidence relevant to the localization result of the region proposal. % a higher score for the region proposal with better localization. 
For the box regression module, we propose a semi-explicit box regressor to strengthen the logic relation between the region feature and its box regression. These logic inferences are category-agnostic, and thus can maintain generalization to the novel category.

Extensive experiments on two large and challenging few-shot detection benchmark datasets, i.e., MS COCO \cite{coco} and FSOD dataset \cite{fsod}, show that IR-FSOD can reach the state-of-the-art few-shot object detection performance while achieving instant response. Especially under the instant response setting, it promotes the current upper-performance limit \cite{fsod} by a large margin.
Our main contributions can be summarized as follows:
\begin{enumerate}
    \item We propose two meta-strategies that can enable the object detection of novel categories with instant response.
    
    \item We further introduce two explicit localization inferences into the localization module in the R-CNN derived model to alleviate its over-fitting to the base categories.

    \item By applying the improvements, we propose a novel Instant Response Few-shot Object Detection (IR-FSOD) framework, which can achieve state-of-the-art results in both response time, precision, and recall.
    
    % \item In the \textbf{Appendix}, we also explore the application advantages of IR-FSOD in various scenarios.
    
\end{enumerate}

\begin{figure*}[t]
    \centering
    \includegraphics[width=1.7\columnwidth]{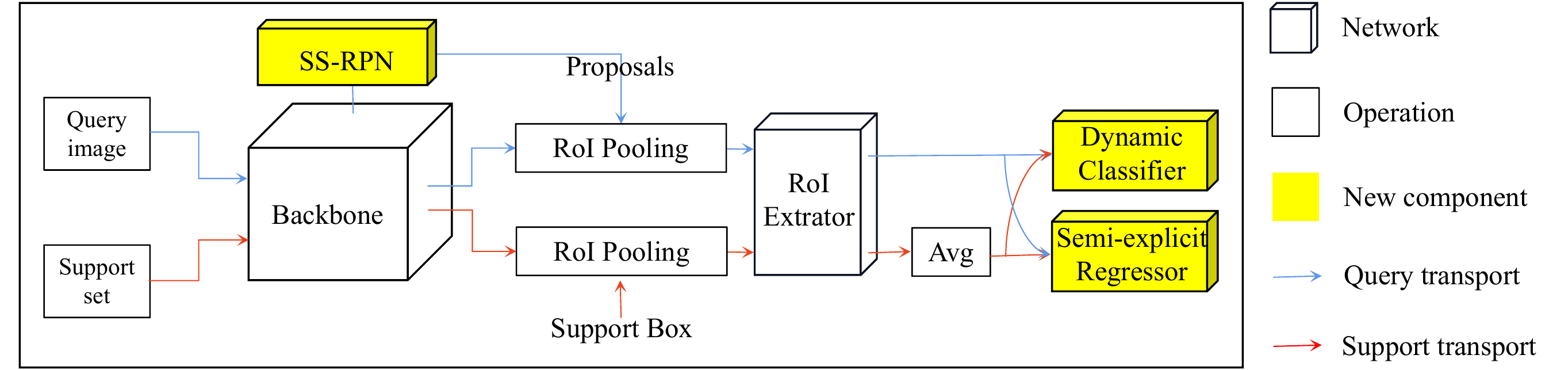}
    \vspace{-10pt}
  \caption{The overview of the IR-FSOD framework. It is built on top of Faster R-CNN. The module without yellow mask is the same as Faster R-CNN. SS-RPN means the proposed semi-supervised RPN. The detail of the semi-supervised RPN, dynamic box classifier, and semi-explicit box regressor are shown in Figure \ref{fig: detail}, Sec. \ref{sec: meta}, and Sec. \ref{sec: framework}.
    }
    \vspace{-10pt}
\label{fig: overview}
\end{figure*}

\section{Related Work}

% \subsection{General Object Detection.}
\noindent\textbf{General Object Detection}
is a fundamental task in computer vision that has attracted lots of attention. Modern object detectors can be divided into two kinds: one-stage detectors and two-stage detectors. One-stage detectors directly predict categories and locations of objects, e.g., YOLO series \cite{yolov1, yolov2, yolov3, yolov4}, SSD \cite{ssd}, etc. Two-stage detectors, pioneered by R-CNN \cite{rcnn}, first generate class-agnostic region proposals, then further refine and classify the proposals \cite{f-rcnn, fast-rcnn, m-rcnn}. These works heavily rely on a huge amount of annotated data and are invalidated on the data with unseen categories, thus they can not be used to solve the FSOD problem.

\noindent\textbf{Few-shot learning} aims to recognize novel classes given limited labeled data. Meta-learning methods \cite{maml, mtl, ml-lstm, reptile, adp_rew}, aka "learning to learn", propose to learn a meta-learner that can fast adapt to new tasks with few labeled samples. Distance metric learning methods \cite{protonet, siamese, relationnet, bipath, tcgfsl} focus on designing a distance formulation between the samples in an embedding space generated by deep neural networks. Popular metrics include cosine similarity \cite{deepface, cosface, closer}, Euclidean distance \cite{protonet} and graph distance \cite{few-gnn}.

% \subsection{Few-shot Object Detection.}
\noindent\textbf{Few-shot Object Detection}
was proposed to handle the object detection for the novel category by only a few annotated samples. There are mainly two types of methods aiming to address the few-shot object detection problem,i.e., meta-learning-based methods and fine-tuning-based methods:

Meta-learning-based methods attempt to build their few-shot detectors by leveraging various meta-learning techniques to extract the class-agnostic knowledge or transfer the knowledge from base categories to novel categories. Despite this goal, these methods still require a fine-tuning process. Otherwise, they are either invalidated \cite{fsrw, meta-rcnn, metar-cnn, gendet, metadet, fsiw, DCNet} or lagging behind other methods \cite{repmet, fsod, dual}. 
FSRW \cite{fsrw} extracts generic meta-features from base categories, then adjusts them using the re-weighting features for novel categories. Meta R-CNN \cite{metar-cnn} and FSIW \cite{fsiw} propose to use the re-weighting features over RoI features instead of the image feature. MetaDet \cite{metadet} and GenDet \cite{gendet} propose to estimate the new parameters in the detector for detecting novel category instances. RepMet \cite{repmet} incorporates distance metric learning into few-shot detection to help classify the proposals. % FSIW \cite{fsiw} performs a similar process as Meta R-CNN \cite{metar-cnn}, but it uses a more complex feature aggregation. 
A-RPN \cite{fsod} and meta-rcnn \cite{meta-rcnn} propose attention-RPN to generate the class-specific region proposal. A-RPN also proposes a multi-relation detector and a contrastive training strategy. Recent DCNet \cite{DCNet} proposes to fully exploit local information to benefit the detection process and alleviate the scale variation problem by context-aware feature aggregation. 

Fine-tuning-based methods adopt the general object detectors and focus on improving the fine-tuning process on the support data to effectively transfer the category-specific model to the novel category. They are once suffered from poor performance, but recent works set the new state-of-the-art. TFA \cite{tfa} simply fine-tunes the last layer of Faster R-CNN \cite{f-rcnn} and substantially improves the performance. MPSR \cite{mpsr} proposes to handle the scale variance issue by multi-scale positive sample refinement, but it needs a manual selection. Recent FSCE \cite{fsce} builds a strong baseline upon TFA \cite{tfa} and boosts the performance by large margins. It also integrates contrastive learning and achieves impressive performance. 
%%
% Although the existing meta-learning-based methods often also require the fine-tuning process, the fine-tuning-based methods require a much longer time.

% As can been seen, fine-tuning is essential for the existing method, no matter what type. Figure \ref{fig: tuning_time} shows the time in some existing methods \cite{tfa, fsod, fsiw, fsce} required for fine-tuning. Note that it requires at least 15 minutes or even a few hours, which is unacceptable in few-shot learning. By comparison, the fine-tuning process in few-shot image classification often requires less than one second. Thus for the application of few-shot object detection, it is valuable to get rid of fine-tuning or replace it with a time-acceptable tuning method.
% adopt the tuning process in an acceptable time.

%%%%%%%%%%%%%%%%%%%%%%%%%%%%%%%%%%%%%%%%%%%%%

\section{METHODOLOGY}

\subsection{Problem Definition}
\label{sec: knowledge}
Given a base dataset $\mathcal{D}_b$ with annotated instances of the base (seen) category $\mathcal{C}_b$, the objective of few-shot object detection is to train a robust model on $\mathcal{D}_b$ which can be generalized on the novel dataset $\mathcal{D}_n$ with instances of the novel (unseen) category $\mathcal{C}_n$ ($\mathcal{C}_n \cap \mathcal{C}_b = \phi $). For each novel category, there is also a support set $\mathcal{S}$ with a few annotated instances, which are only available during testing. In more detail, N-way K-shot object detection means $\mathcal{C}_n$ contains N categories and each support set contains K annotated instances (usually less than 10), i.e., $\mathcal{S}_c = \{(I_i, b_i), i = 1, ..., K \}$ where $I$ and $b$ denote the support image and the bounding box of the support instances. The novel dataset is also called query set.

\subsection{IR-FSOD Framework}
\label{sec: overview}
The overview of IR-FSOD is shown in Figure \ref{fig: overview}, which is based on Faster R-CNN \cite{f-rcnn}. The general process is as follows: Given a query image and a support set, the goal is to detect the objects belonging to the support category in the query image. Firstly, the backbone extracts the feature maps of the query image and all support images. Then the RPN module predicts the region proposals in the query image. After that, the RoI module, including RoI-pooling and RoI-extractor, extracts the feature maps of region proposals and the feature maps of all the support instances, with the same shape. Finally, the box classifier and box regressor further predict the category and the box regression of the region proposals by comparing the region features and the support feature which is the average of all support instance feature maps. In addition to the framework, there is also a non-maximum suppression (NMS) algorithm to select the non-overlap prediction box according to the localization score.

In particular, we first propose two meta-strategies (Sec. \ref{sec: meta}), i.e., dynamic classifier and semi-supervised RPN (SS-RPN), to enable the object detection of novel categories with instant response. Then, we further present two explicit localization inferences (Sec. \ref{sec: framework}), i.e., explicit localization score and semi-explicit box regression, to alleviate the over-fitting to the base categories. As shown in Figure \ref{fig: overview}, we mark the position of the proposed techniques in yellow.

\subsection{Meta-strategies}
\label{sec: meta}
In this section, we focus on the classification module in the framework, including the box classifier and the RPN module.
% The key to across-category task lies in the generalization to the novel category. Therefore, we propose two meta-strategies for the box classifier and RPN module, respectively.

% \subsubsection{Dynamic Classifier}
\noindent\textbf{Dynamic Classifier:} 
As analyzed in Sec. \ref{sec: introduction} and Figure \ref{fig: motivation_dyn}, we argue that the three existing commonly used classifiers all have some defects in the few-shot object detection task. Therefore, we propose to adopt a dynamic classifier module. Specifically, the box classifier module uses different classifiers in training and inference to improve both the generalization to novel categories and the learnability for feature space. In addition, since the distance-classifier is non-parameter, it doesn't require re-training when replacing the trained classifier with the distance-classifier during inference. Table \ref{tab: meta_strategy} shows the ablation study of different classifier combinations on the MS COCO dataset under the instant response setting and 10-shot one-time FSOD evaluation protocol. For a fair comparison, both the multi-classifier and comparison-classifier are single connection layers, and the distance-classifier is the cosine distance between the region feature and the support feature.

As shown in Table \ref{tab: meta_strategy}, the multi-classifier is invalidated on the novel category before re-training. The comparison-classifier performs worse than the distance-classifier since its parameters are still affected by the bias of the base categories. However, the distance-classifier can significantly benefit from models trained by the multi-classifier or comparison-classifier due to their learnability. Based on the results, the IR-FSOD framework adopts the comparison-classifier in the base training and replaces it with the distance-classifier during inference. The details of the two classifiers will be described respectively in Sec. \ref{sec: framework}.

\begin{table}[t]
    \centering
    \small
    \renewcommand\arraystretch{1.3}
    \setlength{\abovecaptionskip}{0.0cm}
    \setlength{\abovecaptionskip}{0.0cm}
    \setlength{\belowcaptionskip}{0.1cm}
% 	\vspace{-5pt}
    \setlength{\tabcolsep}{1.0mm}{
        \caption{ Ablation experimental results for dynamic classifier and semi-supervised rpn on MS COCO under 10-shot setting. }
        
    	\vspace{3pt}
    	
        \label{tab: meta_strategy}
        \begin{tabular}{C{2.0cm}C{2.0cm}|C{0.75cm}C{0.75cm}C{0.75cm}}
        \toprule
        
        \multicolumn{2}{c|}{Dynamic Classifier} & \multirow{2}{*}{$AP$} & \multirow{2}{*}{$AP_{50}$} & \multirow{2}{*}{$AP_{75}$}  \\
        % \cline{4-14}
        (Training) & (Inference)  \\
        \midrule 
        Multi      & Multi     & \multicolumn{3}{c}{invalidated} \\
        Comparison  & Comparison & 4.71 & 8.79  & 4.41 \\
        Distance    & Distance   & 5.94 & 15.64 & 2.82 \\
        Multi      & Distance   & 6.89 & 15.28 & 5.44 \\
        Comparison  & Distance   & \textbf{8.69} & \textbf{17.38} & \textbf{7.78} \\
        \midrule 
        \multicolumn{2}{c|}{ +\ \ Semi-supervised RPN} & \textbf{10.54} & \textbf{20.96} & \textbf{9.08} \\
        
        \bottomrule
        \cr 
        \end{tabular}
        }
	\vspace{-25pt}
\end{table}

\noindent\textbf{Semi-supervised RPN (SS-RPN):} 
Generating region proposals by the class-agnostic detector (e.g., RPN) is a crucial idea in two-stage detection models, but it has a fatal defect in few-shot object detection. As shown in Figure \ref{fig: motivation_rpn}, the positive regions in the training phase are disjoint with the related regions for the novel categories so that these related regions are either ignored or regarded as background in the base training. Thus the RPN module in the FSOD framework is implicitly class-specific to the base categories, which is hard to capture the anchors related to novel categories. 

To address the problem, we propose to adopt semi-supervised algorithms to train the RPN module. Concretely, in the RPN training, all the positive anchors are certain foreground instances. But the negative anchors actually consist of background or potential object instances not belonging to the base categories, so we remark them as unlabeled data. In the IR-FSOD framework, we adopt a simple but effective semi-supervised algorithm, i.e., Pseudo Label.

\begin{figure}[h]
    % \vspace{-10pt}
    \centering
    \includegraphics[width=0.9\columnwidth]{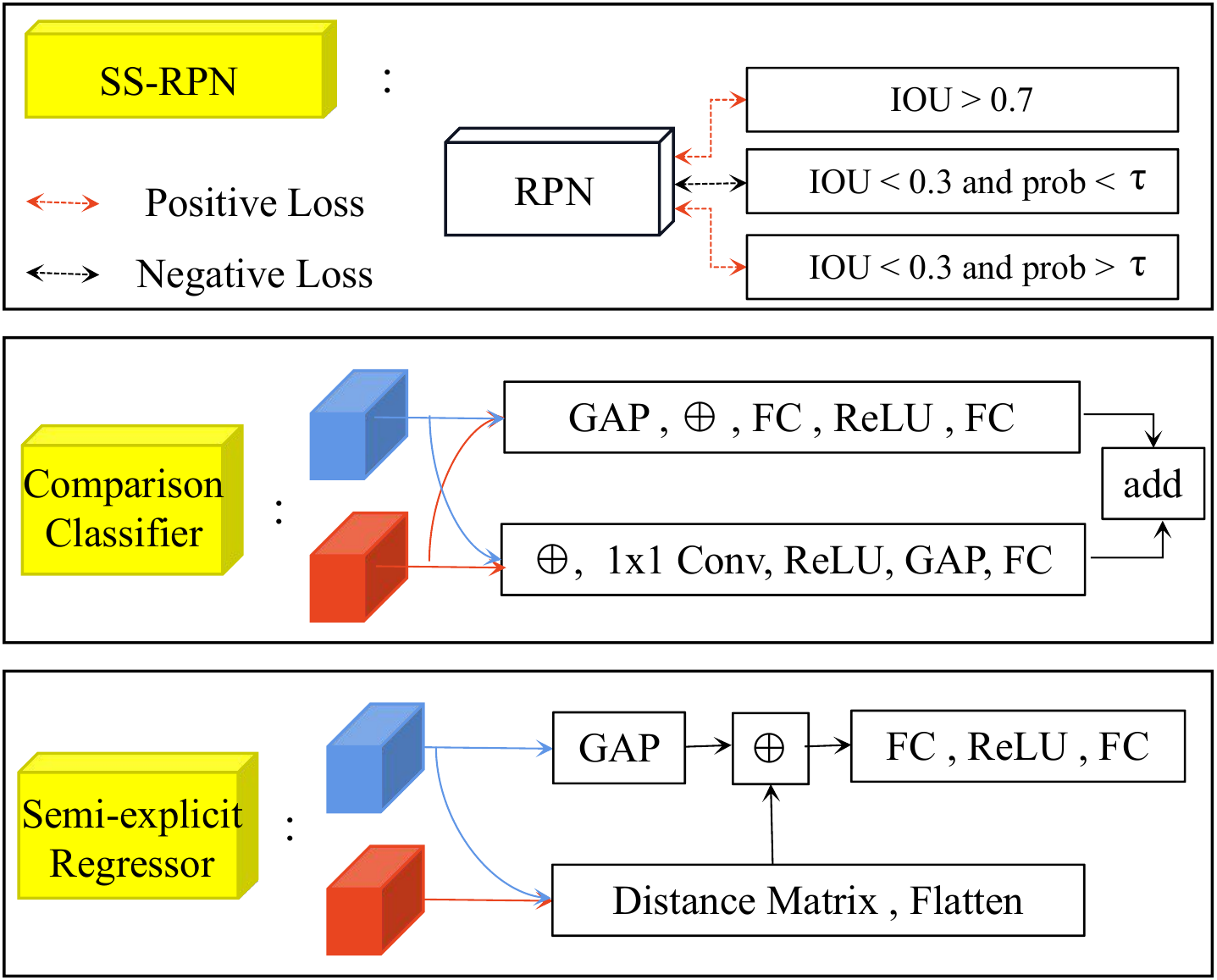}
    \vspace{-5pt}
  \caption{The detail of components in the IR-FSOD. The input of the comparison-classifier and semi-explicit box regressor are the feature map of a region proposal (blue) and the support feature map of a category (red). GAP, FC, and Conv mean the global average pooling, full connection layer, and convolution layer. $\oplus$ means to concatenate the input features.
    }
    \vspace{-15pt}
\label{fig: detail}
\end{figure}

In more detail, as shown at the top of Figure \ref{fig: detail}, we first annotate the anchors whose Intersection over Union (IoU) with the ground truth box is less than 0.3 as negative, and the anchors whose IoU is greater than 0.7 as positive, following the standard RPN training process. Then, we annotate the negative anchors with RPN prediction probabilities greater than threshold $\tau$ as the pseudo positive label and compute positive loss, the same as the positive label. To calculate the balance loss, we keep the ratio of the positive anchors, the negative anchors, and the pseudo positive anchors as 1:1:1. The selection of threshold $\tau$ is shown in Sec. \ref{sec: hyper}.

\textbf{Discussion:} In the object detection task, it is unrealistic to achieve both high recall and high precision. The RPN with semi-supervised training inevitably leads to more background proposals in the inference. However, we argue that this is worth it because it is possible to eliminate these background proposals by the box classifier in the two-stage detection model. On the contrary, the foreground anchors ignored by the RPN module are irreparable. Experiments also validate that even the naive semi-supervised algorithm (i.e., pseudo label) can significantly boost the object detection performance for the novel category (As shown in Table \ref{tab: meta_strategy}).

\subsection{Explicit Localization Inferences}
\label{sec: framework}
In this section, we focus on the localization module in the framework, including the localization score and the box regression.

\begin{figure}[t]
    \centering
    \includegraphics[width=0.9\columnwidth]{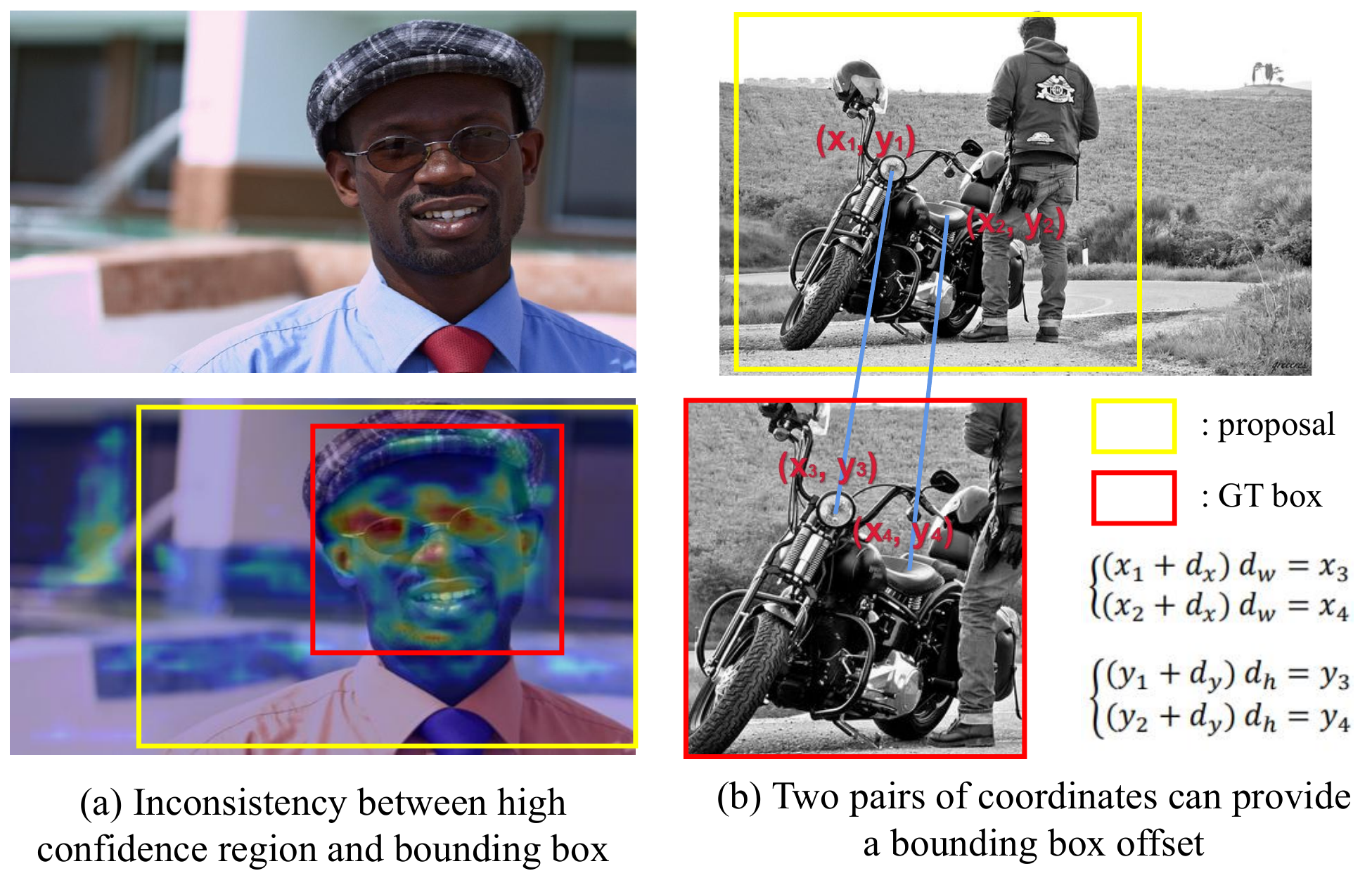}
    \vspace{-5pt}
    \caption{ The motivations of explicit localization inferences. The yellow box and the red box in (a) mean the bounding box and the local high confidence region respective. (b) shows an explicit box regression method.
    }
    \vspace{-10pt}
\label{fig: motivation_spa}
\end{figure}

\noindent\textbf{Explicit Localization Score:} 
R-CNN derived model implicitly evaluates the localization score of the region proposal by the classification score from the box classifier. 
However, the object bounding box usually contains some low-confidence regions, such as the background and the low-discrimination parts of the target object. In contrast, the classification score often only considers the high-confidence region. For example, the classification scores of the two region box in Figure \ref{fig: motivation_spa} (a) are almost the same. Thus the classification score is inconsistent with the localization score, i.e., it cannot reflect the localization result. Although this inconsistency can be alleviated by training on a large amount of annotated data, it can not be generalized to the novel category. Thus it is necessary to explore a training-independent alleviating method. 

To tackle the problem, we propose to integrate the pixel-wise contrast between feature maps into the box classification to explicitly evaluate the localization confidence of the region proposal. Specifically, the RoI module first extracts the feature maps of the region proposal and the support feature with the same shape. Then the core idea is to integrate the features comparison on each pixel of the feature map, which can compare the similarity between the instance distribution in the region proposal and the standard distribution in the support box. Obviously, the higher similarity between the instance distribution can often indicate better localization. We also provide the specific case in the \textbf{Appendix} to illustrate the effectiveness of this approach. As mentioned above, IR-FSOD adopts a dynamic classifier module, thus we design different integration methods for the comparison-classifier and the distance-classifier, respectively. The integration method for the comparison-classifier is shown in the middle of Figure \ref{fig: detail}, which integrates the pixel-wise contrast by a lightweight network.
% As shown in the middle of Figure \ref{fig: detail}, the comparison-classifier integrates the pixel-wise contrast by a lightweight network.

For the distance-classifier, it can integrate the pixel-wise contrast by the distance between the flattened feature maps. Concretely, given a region proposal $x$ and a support set of category $c$, it first calculates the cosine distance between global feature vectors and the cosine distance between flattened feature maps, then adopts their weighted sum to integrate the two distances.
% calculate the prediction score by the sharp sigmoid function $\sigma$. 
Finally, the probability of $x$ belonging to category $c$ is predicted as:

\begin{equation}
        Pr(c;x) = \sigma [\ (1-\alpha)D(f_x, f_c) + \alpha D(v_x, v_c))\ ] \label{eq1}
\end{equation}

\begin{equation}
        D(x, y) = \frac{x^Ty}{||x||\cdot||y||}
\end{equation}

\begin{equation}
        \sigma(x) = \frac{1}{1+e^{-\lambda x}}
\end{equation}

\noindent where $v$ and $f$ represent the vectors obtained from the feature map by global average pooling and flatten function, respectively. $D$ and $\sigma$ mean the cosine distance and sharp sigmoid function. The selection of $\alpha$ and $\lambda$ is shown in Sec. \ref{sec: hyper}.

\noindent\textbf{Semi-explicit Box Regression:}
General object detectors \cite{f-rcnn, yolov1} often implicitly fit the mapping between the features and the box regression by a lightweight network. These mappings are dependent on the category, thus the trained regressor is hard to generalize to the novel category. To tackle the problem, we propose a semi-explicit box regressor by introducing the category-agnostic logic relation into the regression mapping. It leverages an explicit regression mechanism that: any two pairs of coordinates between the region proposal and the GT box can provide two regression equations, equivalent to a correct box regression, as shown in Figure \ref{fig: motivation_spa} (b). Despite the equivalence, this explicit regression is invalidated since the GT box is unavailable during inference. Therefore, we propose to extract sufficient possible coordinate pairs from the comparison between the region proposal and the support box, and then predict the box regression by these coordinate pairs.

% we propose to extract and evaluate the coordinate pairs by the pixel-matching contrast.
Specifically, given the feature map of a region proposal $x$ and the average support feature map of category $c$ (e.g., $ F_x, F_c \in R^{d\times r\times r}$), we first reshape them as list of feature vectors (e.g., $\hat{F_x}, \hat{F_c} \in R^{r^2 \times d}$), then compute the distance matrix $M$ between two lists, where
\begin{equation}
    M_{i, j} = D(\hat{F_x^i}, \hat{F_c^j}).
\end{equation}
Then, we flatten the distance matrix to a distance vector $ d_M \in R^{r^4}$ and concatenate it with the region proposal feature. Finally, we feed the concatenated feature into a lightweight network to predict the box regression, as shown at the bottom of Figure \ref{fig: detail}.

\textbf{Discussion:} In $d_M$, each index represents a coordinate pair between two feature maps, and the corresponding value indicates the confidence score of the coordinate pair. However, these confidence scores may not be accurate due to the difference between the support instance and the GT instance. Explicitly calculating box regression by these coordinate pairs will suffer from serious errors by inaccurate scores. Thus we still predict the box regression by feeding all confidence scores into a neural network to implicitly synthesize all the equations. This regression method is between the implicit regression in general object detectors and the explicit regression by the regression equation, so we call it semi-explicit.

\subsection{Training Strategy:}
\label{sec: training}
Inspired by A-RPN \cite{fsod}, we train our model by the 2-way 10-shot contrastive training strategy.
For each training image as a query image, we first randomly select a positive category $c_1$ that appears in the image and a negative category $c_2$ that doesn't appear in the image ($c_1, c_2 \in \mathcal{C}_b$) and then collect their support sets ($\mathcal{S}_{c_1}$ and $\mathcal{S}_{c_2}$) from $\mathcal{D}_b$, both containing ten object instances.
After the forward process described in \ref{sec: overview}, we train the box comparison-classifier by the positive loss that matches the same category and the negative loss that distinguishes the different categories. For the box regressor, we only calculate the box regression loss of the region proposal belonging to $c_1$.
The training of the semi-supervised RPN is the same as Faster R-CNN \cite{f-rcnn} except for the pseudo-label described above.
The final loss function is defined as:
\begin{equation}
    \mathcal{L} = \mathcal{L}_{rpn} + \mathcal{L}_{cls} + \mathcal{L}_{reg}
\end{equation}
where $\mathcal{L}_{rpn}$ consists of the classification loss and regression loss of proposals, $\mathcal{L}_{cls}$ is the binary cross-entropy loss for box classification, $\mathcal{L}_{reg}$ is the smoothed $L_1$ loss for box regression.

\begin{table*}[h]
    \centering
    \small
    \renewcommand\arraystretch{1.3}
    \setlength{\abovecaptionskip}{0.0cm}
    \setlength{\abovecaptionskip}{0.0cm}
    \setlength{\belowcaptionskip}{0.1cm}
    \setlength{\tabcolsep}{1.0mm}{
        \caption{ Few-shot detection results for 20 novel classes on COCO dataset. ``IR'' means the model is instant response, i.e., without tuning process. \textcolor{red}{RED}/\textcolor{blue}{BLUE} indicate the SOTA/second best. $^+$ means the result is estimated by the description in their paper.}
        \vspace{3pt}
        \label{tab: coco_10shot}
        \begin{tabular}{cc|C{0.84cm}C{0.84cm}C{0.84cm}C{0.84cm}C{0.84cm}C{0.84cm}|C{0.84cm}C{0.84cm}cC{0.84cm}C{0.84cm}C{0.84cm}|c}
        \toprule
        \multirow{2}{*}{Model} & \multirow{2}{*}{Backbone} & \multicolumn{6}{c|}{Average Precision}  & \multicolumn{6}{c|}{Average Recall} & Tuning\\
        % \cline{4-14}
        & & $AP$ & $AP_{50}$ & $AP_{75}$ & $AP_{S}$ & $AP_{M}$ & $AP_{L}$ & $AR_1$ & $AP_{10}$ & $AR_{100}$ & $AR_S$ & $AR_M$ & $AR_L$ & time \\
        \midrule 
        % \multirow{1}{*}{TFSL} 
        A-RPN   \cite{fsod}     & Res-50  & 7.3 & 13.2 & 7.1 & \textcolor{blue}{\textbf{4.4}} & 8.7 & 10.7 & 17.5 & \textcolor{blue}{\textbf{32.3}} & \textcolor{blue}{\textbf{33.2}} & \textcolor{blue}{\textbf{10.0}} & \textcolor{blue}{\textbf{34.7}} & \textcolor{blue}{\textbf{50.4}} & \textbf{IR} \\
        Meta R-CNN \cite{metar-cnn} & Res-50 & 8.7 & 19.1 & 6.6 & 2.3 & 7.7 & 14.0 & 12.6 & 17.8 & 17.9 & 7.8 & 15.6 & 27.2  & 5$^+$ h \\
        MPSR    \cite{mpsr}     & FPN-101 & 9.8 & 17.9 & 9.7 & 3.3 & 9.2 & 16.1 & 15.7 & 21.2 & 21.2 & 4.6 & 19.6 & 34.3 & 40$^+$ min \\
        TFA     \cite{tfa}      & FPN-101 & 9.8 & 19.7 & 8.9 & 2.8 & 9.2 & 16.1 & 14.5 & 18.6 & 18.6 & 5.3 & 14.8 & 33.1 & 16 h \\
        FSCE    \cite{fsce}     & FPN-101 & 11.9 & 22.3 & 11.6 & 2.9 & 11.1 & 17.6 & 17.0 & 26.6 & 26.5 & 6.7 & 26.3 & 42.3 & 3 h \\
        A-RPN+FT\cite{fsod}     & Res-50  & 12.0 & 22.4 & \textcolor{blue}{\textbf{11.8}} & 2.9 & 12.2 & 20.7 & 18.8 & 26.4 & 26.4 & 3.6 & 23.6 & 45.6 & 15 min \\
        FSIW    \cite{fsiw}     & Res-50  & 12.5 & \textcolor{red}{\textbf{27.3}} & 9.8 & 2.5 & 13.8 & 19.9 & \textcolor{red}{\textbf{20.0}} & 25.5 & 25.7 & 7.5 & 27.6 & 38.9 & 5 h  \\
        DCNet   \cite{DCNet}    & Res-101 & \textcolor{blue}{\textbf{12.8}} & 23.4 & 11.2 & 4.3 & \textcolor{blue}{\textbf{13.8}} & \textcolor{blue}{\textbf{21.0}} & 18.1 & 26.7 & 25.6 & 7.9 & 24.5 & 36.7 & 40$^+$ min \\
        \midrule
        % IR-FSOD      & Res-50  & 12.5 & 23.6 & 11.6 & \textcolor{red}{\textbf{6.2}} & 15.6 & 21.1 & 19.1 & 33.5 & 35.6 & 11.4 & 39.4 & 54.9 & \textbf{IR} \\
        IR-FSOD & Res-50  & \textcolor{red}{\textbf{13.1}} & \textcolor{blue}{\textbf{24.5}} & \textcolor{red}{\textbf{12.3}} & \textcolor{red}{\textbf{5.9}} & \textcolor{red}{\textbf{16.2}} & \textcolor{red}{\textbf{22.0}} & \textcolor{blue}{\textbf{19.1}} & \textcolor{red}{\textbf{33.5}} & \textcolor{red}{\textbf{35.6}} & \textcolor{red}{\textbf{11.4}} & \textcolor{red}{\textbf{39.4}} & \textcolor{red}{\textbf{54.9}} & \textbf{IR} \\
         % \multirow{1}{*}{TFSL} 
         
         \bottomrule
         \cr 
        \end{tabular}
        }
	\vspace{-21pt}
\end{table*}

\begin{table*}[htb]
    \centering
    \small
    \renewcommand\arraystretch{1.3}
    \setlength{\abovecaptionskip}{0.0cm}
    \setlength{\abovecaptionskip}{0.0cm}
    \setlength{\belowcaptionskip}{0.1cm}
    \setlength{\tabcolsep}{1.0mm}{
        \caption{ Few-shot detection results on COCO dataset under different shot settings. * means the result is re-implemented. \textcolor{red}{RED}/\textcolor{blue}{BLUE} indicate the SOTA/second best. The results are averaged over ten random runs.  }
        \vspace{3pt}
        \label{tab: coco_1-10shot}
        \begin{tabular}{cc|C{0.71cm}C{0.71cm}C{0.71cm}|C{0.71cm}C{0.71cm}C{0.71cm}|C{0.71cm}C{0.71cm}C{0.71cm}|C{0.71cm}C{0.71cm}C{0.71cm}|C{0.71cm}C{0.71cm}C{0.71cm}}
        \toprule
        \multirow{2}{*}{Model} & \multirow{2}{*}{Backbone} & \multicolumn{3}{c|}{1-shot}  & \multicolumn{3}{c|}{2-shot} & \multicolumn{3}{c|}{3-shot} & \multicolumn{3}{c|}{5-shot} & \multicolumn{3}{c}{10-shot}  \\
        % \cline{4-14}
        & & $AP$ & $AP_{50}$ & $AP_{75}$ & $AP$ & $AP_{50}$ & $AP_{75}$ & $AP$ & $AP_{50}$ & $AP_{75}$ & $AP$ & $AP_{50}$ & $AP_{75}$ & $AP$ & $AP_{50}$ & $AP_{75}$  \\
        \midrule 
        % \multirow{1}{*}{TFSL} 
        A-RPN* \cite{fsod}     & Res-50  & \textcolor{blue}{\textbf{3.6}} & 7.2 & \textcolor{blue}{\textbf{3.2}} & \textcolor{blue}{\textbf{5.1}} & 9.7 & \textcolor{blue}{4.7} & 5.6 & 10.7 & 5.2 & 6.3 & 11.9 & 5.9 & 6.7 & 12.5 & 5.8 \\
        % MPSR    \cite{mpsr}     & FPN-101 &  \\
        TFA     \cite{tfa}      & FPN-101 & 1.9 & 3.8 & 1.7 & 3.9 & 7.8 & 3.6 & 5.1 & 9.9 & 4.8 & 7.0 & 13.3 & 6.5 & 9.1 & 17.1 & 8.8  \\
        FSIW    \cite{fsiw}     & Res-50  & 3.2 & \textcolor{blue}{\textbf{8.9}} & 1.4 & 4.9 & \textcolor{blue}{\textbf{13.3}} & 2.3 & \textcolor{blue}{\textbf{6.7}} & \textcolor{red}{\textbf{18.6}} & 2.9 & \textcolor{blue}{\textbf{8.1}} & \textcolor{blue}{\textbf{20.1}} & 4.4 & 10.7 & \textcolor{red}{\textbf{25.6}} & 6.5  \\
        A-RPN+FT*\cite{fsod}     & Res-50  & - & - & - & 4.8 & 9.2 & 3.9 & 5.9 & 11.6 & \textcolor{blue}{\textbf{5.7}} & \textcolor{blue}{\textbf{8.1}} & 15.5 & \textcolor{blue}{\textbf{7.4}} & 10.9 & 20.5 & 9.4   \\
        FSCE*    \cite{fsce}     & FPN-101 & 2.0 & 4.9 & 1.3 & 4.2 & 9.5 & 3.4 & 5.7 & 12.0 & 4.7 & 7.6 & 15.6 & 6.5 & \textcolor{blue}{\textbf{11.2}} & 22.3 & \textcolor{blue}{\textbf{9.8}}  \\
        \midrule
        % IR-FSOD    & Res-50  & 5.1 & 10.8 & 4.3 & 7.7 & 15.7 & 6.6 & 8.9 & 17.6 & 7.9 & 10.5 & 20.8 & 9.1 & 12.0 & 23.5 & 10.8  \\
        IR-FSOD & Res-50  & \textcolor{red}{\textbf{5.1}} & \textcolor{red}{\textbf{10.8}} & \textcolor{red}{\textbf{4.3}} & \textcolor{red}{\textbf{7.7}} & \textcolor{red}{\textbf{15.7}} & \textcolor{red}{\textbf{6.6}} & \textcolor{red}{\textbf{8.9}} & \textcolor{blue}{\textbf{17.6}} & \textcolor{red}{\textbf{7.9}} & \textcolor{red}{\textbf{10.5}} & \textcolor{red}{\textbf{20.8}} & \textcolor{red}{\textbf{9.1}} & \textcolor{red}{\textbf{12.0}} & \textcolor{blue}{\textbf{23.5}} & \textcolor{red}{\textbf{10.8}}  \\
         % \multirow{1}{*}{TFSL} 
         
         \bottomrule
         \cr 
        \end{tabular}
        }
	\vspace{-21pt}
\end{table*}

% \clearpage

\section{Experiments}

\subsection{Experimental Setup}
\label{sec: experimentsetup}

\textbf{Dataset:}
In this paper, we conduct experiments on two large and challenging few-shot detection benchmark datasets, MS COCO \cite{coco} and FSOD dataset \cite{fsod}, which contains 800K objects belonging to 80 categories and 182K objects belonging to 1000 categories respectively. For MS COCO, we set the 20 categories belonging to PASCAL VOC \cite{voc} as the novel categories and the remaining 60 categories as the base categories following the existing works \cite{tfa, fsce, fsod, fsrw}. We use the \textit{train2017} with only annotations of base categories for training and evaluate the detection result of novel categories on the \textit{val2017}. FSOD dataset is specially designed for few-shot object detection, whose training set and test set only contain disjoint 800 base categories and 200 novel categories.
%%
% We perform the same evaluation process of MS COCO on the FSOD test set.

\noindent\textbf{Implement Details:}
We use the commonly used ResNet-50 \cite{resnet} as our backbone. Following the existing works \cite{fsod, tfa, fsce}, the backbone is pre-trained on ImageNet \cite{imagenet}. The most of network architectures and the hyper-parameters remain the same as Faster R-CNN \cite{f-rcnn} except for the box classifier and box regressor, as described in Sec. \ref{sec: framework}. In addition, we halve the number of sampled anchors in RPN and proposals in the RoI head used for loss calculation from (512, 256) for the positive and negative anchors to (128, 128, 128) for the positive, the negative, and the pseudo positive anchors during training. Our model is trained by SGD optimizer on 3 RTX 2080Ti GPUs with a batch size of 9 (3 query images per GPU) for 120,000 iterations. The learning rate is initialized as 0.003 with the weight decay of factor 0.1 at 80,000$^{th}$ and 110,000$^{th}$ iteration.
%%
% Following A-RPN \cite{fsod}, we pre-collect all the support set by cropping, zero-padding, and resizing the support instances to $320\times 320$ to reduce the computation cost of the support set.

\noindent\textbf{Evaluation protocol:}
We conduct experiments on the one-time FSOD protocol proposed in \cite{fsrw} and the meta-testing protocol commonly used in the few-shot learning \cite{matchnet}. Given the support sets of all novel categories, the one-time FSOD protocol directly evaluates the performance of detecting these novel categories on the complete test set. The meta-testing protocol requires evaluating the average performance of the detector under numerous random episodes. Each episode randomly collects a novel category subset and consists of the corresponding support set and query set. 
In addition, we also provide the experiments under the class-incomplete setting in the \textbf{Appendix} since it is rarely used.

\subsection{One-time FSOD evaluation}
\label{sec: sota}
\noindent\textbf{MS COCO result: } In Table \ref{tab: coco_10shot}, we compare our IR-FSOD with the previous state-of-the-art methods under the 10-shot setting. For a fair comparison, we also report the backbone used in the models and the time required for the tuning process. Although the competitive DCNet \cite{DCNet} doesn't adopt FPN, it still uses multi-scale features to enhance its detector. As shown in the table, IR-FSOD achieves new state-of-the-art results in most indicators. Under the instant response setting, it outperforms the latest method \cite{fsod} by about 78$\%$ on the $AP$ metric. It's worth noting that IR-FSOD achieves both the highest precision and recall, which is rare in the existing methods. For example, A-RPN \cite{fsod} before fine-tuning is competitive to our model on the recall, but its precision ($AP$) is only 56$\%$ of ours; DCNet \cite{DCNet} is competitive to our model on the precision, but its recall ($AR_{10}$) is only 80$\%$ of ours. To sum up, IR-FSOD achieves the overall leadership in response time, precision, and recall.

Then, we conduct further comparison experiments under different shot settings ($K \in \{1, 2, 3, 5, 10\}$). For a fair comparison, we evaluate all the methods over ten random runs. In each run, all the methods adopt the same support set. The support sets are generated from TFA \cite{tfa}. Besides IR-FSOD, A-RPN \cite{fsod} is the only method that can support instant response, and others require fine-tuning time ranging from ten minutes to several hours per run to respond. As shown in Table \ref{tab: coco_1-10shot}, IR-FSOD can outperform the previous works by 0.8$\%$-2.6$\%$ $AP$ under different shot settings. Although A-RPN can also outperform other methods under the low-shot settings while supporting instant response, it becomes significantly behind the method with fine-tuning as the shot number increase. On the contrary, IR-FSOD can always stay ahead, demonstrating its generalized effectiveness under varied few-shot settings.

\noindent\textbf{FSOD dataset result:} 
Similar to MS COCO, we evaluate all the methods over ten random runs, and all the methods adopt the same support set in each run. The support sets are generated from the code of TFA \cite{tfa}. The average results under the 1/3/5 shot setting are shown in Table \ref{tab: fsod}. As shown in the table, IR-FSOD achieves state-of-the-art results on 1/3 shots and comparable results on 5-shots with instant response, demonstrating the strong generalization to the various novel categories. It should be noted that the FSOD dataset has 200 novel classes, i.e., the support set in the 5-shot setting has 1000 object instances. It's usually hard to obtain so many instances in the practice scenario. Therefore, the detection performance under the low-shot setting is more important. In addition, IR-FSOD also performs both high precision and high recall on the FSOD dataset, the same as the results on the MS COCO dataset, indicating that it is not accidental on a particular dataset.

\begin{table}[h]
	\vspace{-10pt}
    \centering
    \small
    \renewcommand\arraystretch{1.3}
    \setlength{\abovecaptionskip}{0.0cm}
    \setlength{\abovecaptionskip}{0.0cm}
    \setlength{\belowcaptionskip}{0.1cm}
    \setlength{\tabcolsep}{1.0mm}{
        \caption{  Few-shot detection results for 200 novel classes on FSOD dataset. ``Time'' means the tuning time. ``IR'' means the model is instant response, i.e., without tuning process. \textcolor{red}{RED}/\textcolor{blue}{BLUE} indicate the SOTA/second best. All results are re-implemented and averaged over ten random runs. }
        \vspace{3pt}
        \label{tab: fsod}
        \begin{tabular}{ccc|C{0.78cm}C{0.78cm}C{0.78cm}C{0.78cm}c}
        \toprule
        %  \multirow{2}{*}{Method} & \multicolumn{4}{c}{5-way 5-shot} \\
        Shot & Method & Backbone & $AP$ & $AP_{50}$ & $AP_{75}$ & $AR_{10}$ & Time \\ 
        \midrule 
        \multirow{4}{*}{1}
        & A-RPN \cite{fsod}  & Res-50  & \textcolor{blue}{\textbf{9.79}} & \textcolor{blue}{\textbf{16.00}} & \textcolor{blue}{\textbf{10.25}} & \textcolor{red}{\textbf{40.33}} & \textbf{IR}\\
        & TFA  \cite{tfa}   & FPN-101 & 7.43 & 12.07 & 7.79 & 14.42 & 2 h \\ 
        & FSCE \cite{fsce}  & FPN-101 & 7.21 & 12.54 & 6.92 & 15.20 & 1.5 h \\
        & IR-FSOD         & Res-50  &\textcolor{red}{\textbf{10.66}} & \textcolor{red}{\textbf{18.41}} & \textcolor{red}{\textbf{16.62}} & \textcolor{blue}{\textbf{38.47}} & \textbf{IR}\\
            
        \midrule 
        \multirow{4}{*}{3}
        & A-RPN \cite{fsod}  & Res-50  &  14.94 & 23.68 & \textcolor{blue}{\textbf{15.91}}  & \textcolor{red}{\textbf{50.01}} &  \textbf{IR}\\
        & TFA  \cite{tfa}   & FPN-101 & 13.21 &  21.10 &  14.16 & 24.87 & 3 h\\
        & FSCE \cite{fsce}  & FPN-101 & \textcolor{blue}{\textbf{14.96}} & \textcolor{blue}{\textbf{25.85}} & 14.73 & 29.21 & 3 h \\
        & IR-FSOD               & Res-50   & \textcolor{red}{\textbf{16.33}} & \textcolor{red}{\textbf{27.59}} & \textcolor{red}{\textbf{10.65}}  & \textcolor{blue}{\textbf{48.91}} &  \textbf{IR}\\

        \midrule 
        \multirow{4}{*}{5}
        & A-RPN \cite{fsod}  & Res-50  & 17.28 & 27.13 &  18.49  & \textcolor{blue}{\textbf{52.30}} &  \textbf{IR}\\
        & TFA  \cite{tfa}   & FPN-101 &  15.85 & 24.89 & 17.31 & 27.74 & 3.5 h \\
        & FSCE \cite{fsce}  & FPN-101 & \textcolor{red}{\textbf{19.58}} & \textcolor{red}{\textbf{33.42}} & \textcolor{red}{\textbf{19.79}} & 36.10 & 3.5 h \\
        & IR-FSOD            & Res-50  & \textcolor{blue}{\textbf{19.24}} & \textcolor{blue}{\textbf{32.08}} & \textcolor{blue}{\textbf{19.75}}  & \textcolor{red}{\textbf{52.69}} &  \textbf{IR}\\
        
         \bottomrule
        \end{tabular}
        }
	\vspace{-10pt}
    
    \label{tab: episode}
    
\end{table}

\begin{table}[h]
    \centering
    \small
    \renewcommand\arraystretch{1.3}
    \setlength{\abovecaptionskip}{0.0cm}
    \setlength{\abovecaptionskip}{0.0cm}
    \setlength{\belowcaptionskip}{0.1cm}
    \setlength{\tabcolsep}{1.0mm}{
        \caption{ Meta-testing evaluation with 95$\%$ confidence interval on the MS-COCO dataset under 5-way and 1,000 episodes setting. SPE means seconds-per-episode. }
        \vspace{3pt}
        \label{tab: episode_5}
        \begin{tabular}{cc|C{1.5cm}C{1.5cm}C{1.5cm}C{0.6cm}}
        \toprule
        %  \multirow{2}{*}{Method} & \multicolumn{4}{c}{5-way 5-shot} \\
        K & Method & $AP$ & $AP_{50}$ & $AP_{75}$ & SPE \\ 
        
        % \midrule 
        % \multirow{3}{*}{1}
        %  & A-RPN \cite{fsod}  & $9.87_{\pm 0.25}$ & $19.33_{\pm 0.44}$ & $8.94_{\pm 0.27}$ & 9   \\
        %  & IR-FSOD               & $10.36_{\pm 0.27}$ & $21.05_{\pm 0.45}$ & $9.05_{\pm 0.32}$ & \textbf{5} \\ 
        %  & IR-FSOD+CT            & $\textbf{10.78}_{\pm 0.29}$ & $\textbf{21.81}_{\pm 0.50}$ & $\textbf{9.41}_{\pm 0.33}$ & 7\\
        
        \midrule 
        \multirow{3}{*}{5}
         & DANA \cite{dual}  & $12.60_{\pm 0.29}$ & $ 25.90_{\pm 0.44}$ & $ 11.30_{\pm 0.35}$ & 10\\
         & A-RPN \cite{fsod}  & $14.27_{\pm 0.27}$ & $26.61_{\pm 0.45}$ & $13.58_{\pm 0.31}$ & 9\\
         & IR-FSOD\ \ \         & \textbf{16.59}$_{\pm 0.28}$ & \textbf{31.97}$_{\pm 0.47}$ & \textbf{15.12}$_{\pm 0.33}$ & \textbf{5} \\ 
        \midrule 
        
        \multirow{2}{*}{10}
         & A-RPN \cite{fsod}  & $15.12_{\pm 0.29}$ & $27.74_{\pm 0.47}$ & $14.61_{\pm 0.32}$ & 10\\
         & IR-FSOD\ \ \         & \textbf{17.74}$_{\pm 0.29}$ & \textbf{33.59}$_{\pm 0.47}$ & \textbf{16.56}$_{\pm 0.34}$ & \textbf{5} \\
        
         \bottomrule
        \end{tabular}
        }
	\vspace{-12pt}

\end{table}

\begin{table}[h]
    \centering
    \small
    \renewcommand\arraystretch{1.3}
    \setlength{\abovecaptionskip}{0.0cm}
    \setlength{\abovecaptionskip}{0.0cm}
    \setlength{\belowcaptionskip}{0.1cm}
    \setlength{\tabcolsep}{1.0mm}{
        \caption{ Meta-testing evaluation with 95$\%$ confidence interval on the MS-COCO dataset under 10-way and 1,000 episodes setting. SPE means seconds-per-episode.}
        \vspace{3pt}
        \label{tab: episode_10}
        \begin{tabular}{cc|C{1.5cm}C{1.5cm}C{1.5cm}C{0.6cm}}
        \toprule
        %  \multirow{2}{*}{Method} & \multicolumn{4}{c}{5-way 5-shot} \\
        K & Method & $AP$ & $AP_{50}$ & $AP_{75}$ & SPE \\ 
        
        \midrule 
        \multirow{2}{*}{5}
         & A-RPN \cite{fsod}  & $11.32_{\pm 0.19}$ & $20.84_{\pm 0.28}$ & $10.87_{\pm 0.20}$ & 28\\
         & IR-FSOD\ \ \       & \textbf{14.23}$_{\pm 0.16}$ & \textbf{27.39}$_{\pm 0.26}$ & \textbf{13.11}$_{\pm 0.18}$ & \textbf{12}\\
        \midrule 
        
        \multirow{2}{*}{10}
         & A-RPN \cite{fsod}  & $12.11_{\pm 0.15}$ & $22.05_{\pm 0.27}$ & $11.78_{\pm 0.19}$ & 30\\
         & IR-FSOD\ \ \         & \textbf{15.52}$_{\pm 0.14}$ & \textbf{29.41}$_{\pm 0.26}$ & \textbf{14.55}$_{\pm 0.16}$ & \textbf{13}\\
        
         \bottomrule
        \end{tabular}
        }
	\vspace{-12pt}

\end{table}

\subsection{Meta-testing protocol}
\label{sec: meta-test}
In this section, we perform the meta-testing protocol on the MS COCO dataset. For an $N$-way $K$-shot few-shot object detection, we collect 1,000 episodes and evaluate the average object detection performance with 95$\%$ confidence interval. Each episode consists of an $N$-way $K$-shot support set and a query set containing ten images for each category. Since the evaluation is performed on each episode independently, including the fine-tuning process and the inference process, the models with fine-tuning require unacceptable time. For example, DCNet \cite{DCNet} requires more than a month to perform the whole meta-testing. Therefore, we only compare our IR-FSOD with the models that support instant response \cite{fsod, dual}.

Table \ref{tab: episode_5} and \ref{tab: episode_10} report the average results with the 95 $\%$ confidence interval and the detection time (seconds-per-episode) under different few-shot settings, including $K \in \{5, 10 \}$ and $N \in \{5, 10 \}$. As shown in the table, our IR-FSOD achieves a significant lead in both performance and efficiency. Specifically, it outperform A-RPN by 2.3$\%$-3.4$\%$ $AP$, 5.4$\%$-7.3$\%$ $AP_{50}$, and 1.5$\%$-2.8$\%$ $AP_{75}$. In addition, IR-FSOD runs only half as long as A-RPN, since A-RPN introduces many complexities, such as generating the class-specific proposal for each category and integrating multiple relation modules. In contrast, the approaches in IR-FSOD are simpler but more effective.
% indicating that our approaches are simpler and more effective.

\begin{table}[h]
	\vspace{-6pt}
    \centering
    \small
    \renewcommand\arraystretch{1.3}
    \setlength{\abovecaptionskip}{0.0cm}
    \setlength{\abovecaptionskip}{0.0cm}
    \setlength{\belowcaptionskip}{0.1cm}
    \setlength{\tabcolsep}{1.2mm}{
        \caption{  Ablation for key components proposed in this paper: results from on the COCO dataset under the 10-shot setting. }
        \vspace{3pt}
        \label{tab: ablation}
        \begin{tabular}{llcc|C{0.75cm}C{0.75cm}C{0.75cm}}
        \toprule
        %  \multirow{2}{*}{Method} & \multicolumn{4}{c}{5-way 5-shot} \\
        \multicolumn{4}{c|}{Ablation} & $AP$ & $AP_{50}$ & $AP_{75}$ \\ 
        \midrule 
        \multirow{3}{*}{\makecell[c]{Faster R-CNN \\ (baseline)}} & \multicolumn{3}{c|}{ + Multi-classifier} & \multicolumn{3}{c}{invalidated} \\
        & \multicolumn{3}{c|}{ + Comparison-classifier} & 4.71 & 8.79  & 4.41 \\
        & \multicolumn{3}{c|}{ + Distance-classifier} & 5.94 & 15.64 & 2.82 \\
        \midrule 
        \multirow{2}{*}{+ Dyn-cls: }
        % & (Comparison & + & Comparison) & 4.71 & 8.79  & 4.41\\
        % & (Distance & + & Distance) & 5.94 & 15.64 & 2.82\\
        & (Multi & + & Distance) & 6.89 & 15.28 & 5.44\\
        & (Comparison & + & Distance) & \textbf{8.69} & \textbf{17.38} & \textbf{7.78}\\
        \midrule 
        { + SS-RPN} & & & & 10.54 & 20.96 & 9.08 \\
        { + SE-Reg} & & & & 10.64 & 20.59 & 9.58 \\
        { + Cls-PW} & & & & 10.67 & 20.92 & 9.47 \\
        { + SS-RPN} & \multicolumn{3}{l|}{+ Cls-PW} & 11.65 & 22.52 & 10.31 \\
        { + SS-RPN} & \multicolumn{3}{l|}{+ SE-Reg} & 11.82 & 22.92 & 10.60 \\
        { + Cls-PW} & \multicolumn{3}{l|}{+ SE-Reg} & 11.94 & 22.21 & 11.19 \\
        { + SS-RPN} & \multicolumn{3}{l|}{+ Cls-PW\ \ \ \ \ \ \ + SE-Reg} & \textbf{13.05} & \textbf{24.50} & \textbf{12.33}  \\

         \bottomrule
        \end{tabular}
        }
	\vspace{-10pt}
\end{table}

\subsection{Ablation Studies}
\label{sec: ablation}
In this section, we evaluate the effects of the core components in IR-FSOD. All ablation studies are conducted on the COCO dataset under the 10-shot setting and one-time FSOD evaluation protocol. IR-FSOD is built on top of Faster R-CNN \cite{rcnn}, which is designed for general object detection. Thus we adopt it as the baseline and design the ablation experiments in two stages in Table \ref{tab: ablation}.

In the first stage, we evaluate the effect of the proposed dynamic classifier module, which is the essential strategy to transform the general object detector into a few-shot object detector with instant response. Without the dynamic classifier module, the model suffers from low performance or even invalidation due to low learnability or low generalization of the classifier. Just by introducing the dynamic classifier module into the Faster R-CNN, it is already comparable with some methods requiring fine-tuning \cite{metar-cnn, metadet, meta-rcnn}. % while provding instant response.

In the second stage, we evaluate the different combinations of three proposed boosted modules, including the semi-supervised RPN (SS-RPN), the box classifier with the pixel-wise contrast (Cls-PW), and the semi-explicit box regressor (SE-Reg). As shown in Table \ref{tab: ablation}, their improvements for the model performance are different and are all in line with our expectations. Concretely, (a) The semi-supervised RPN mainly achieves the performance improvement on the $AP_{50}$ metric (+1.4$\%$-3.6$\%$), indicating that it successfully captured more potential region proposals belonging to the novel categories; (b) The semi-explicit box regressor can significantly improve the result on the $AP_{75}$ metric (+1.6$\%$-2.0$\%$), which shows that it can generate more accurate high-quality boxes by improving the localization accuracy; (c) The pixel-wise contrast in the box classifier mainly improves the confidence ranking of all the predicted boxes, thus can achieve significant improvement on all evaluation metrics, e.g.,  $AP_{50}$ (+1.4$\%$-3.6$\%$) and $AP_{75}$ (+1.3$\%$-1.7$\%$).
% since it can not only improve the classification accuracy but also improve the localization accuracy by generating a higher confidence for the region proposal with better localization.

\begin{figure}[htb]
    \centering
    \vspace{-5pt}
    \includegraphics[width=\columnwidth]{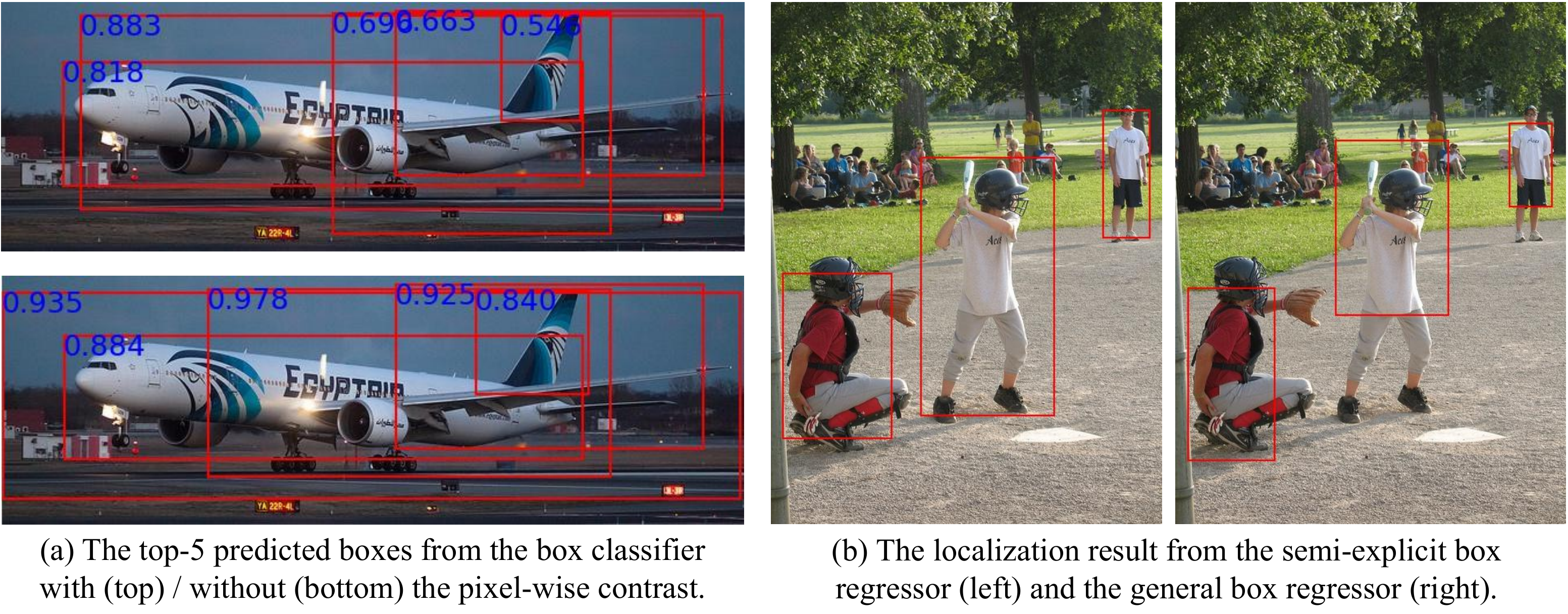}
    \vspace{-20pt}
    \caption{ The qualitative ablation studies for the introduced explicit localization inference, i.e., (a) the pixel-wise contrast and (b) the semi-explicit box regressor.
    }
    % \vspace{-10pt}
\label{fig: qualitative}
\end{figure}

Finally, we also perform qualitative ablation studies for the two explicit localization inferences to verify their improvement in the localization process. As shown in Figure \ref{fig: qualitative} (a), the classifier with the pixel-wise contrast can produce the confidence (top) that is basically consistent with the localization result (i.e., the IoU between ground truth). Without the pixel-wise contrast, the confidence (bottom) is irrelevant to localization, resulting in high-confidence but poorly localized predictions. Figure \ref{fig: qualitative} (b) shows that the semi-explicit box regressor can significantly improve the localization accuracy and generate more high-quality predicted boxes. We also provide more qualitative ablation studies in the \textbf{Appendix} by more detection cases and the statistics of detection results.
% specific detection cases for the qualitative ablation studies in the \textbf{Appendix}.d

\begin{table}[t]
    \centering
    \small
    \renewcommand\arraystretch{1.3}
    \setlength{\abovecaptionskip}{0.0cm}
    \setlength{\abovecaptionskip}{0.0cm}
    \setlength{\belowcaptionskip}{0.1cm}
    \setlength{\tabcolsep}{1.0mm}{
        \caption{  Hyper-parameter studies of $\tau$ and $\alpha$ in the IR-FSOD: results from on the COCO dataset under the 10-shot setting. }
        \vspace{3pt}
        \label{tab: param}
        \begin{tabular}{C{0.6cm}|C{0.8cm}C{0.8cm}C{0.8cm}C{0.2cm}C{0.6cm}|C{0.8cm}C{0.8cm}C{0.8cm}}
        % \toprule
        \cline{1-4} \cline{6-9} 
        $\tau$ & $AP$ & $AP_{50}$ & $AP_{75}$ & & $\alpha$ & $AP$ & $AP_{50}$ & $AP_{75}$ \\ 
        % \midrule 
        \cline{1-4} \cline{6-9} 
        
        -    & 11.94 & 22.21 & 11.19 & & 0.0 & 11.82 & 22.92 & 10.60 \\
        0.75 & 12.38 & 22.98 & 11.73 & & 1/3 & 12.96 & 24.39 & 12.16 \\
        0.50 & 12.66 & 23.97 & 11.79 & & 1/2 & \textbf{13.05} & \textbf{24.50} & \textbf{12.33} \\
        0.25 & \textbf{13.05} & \textbf{24.50} & \textbf{12.33} & & 2/3 & 12.92 & 24.20 & 12.18 \\
        0.10 & 12.87 & 24.05 & 12.03  & & 1.0 & 12.01 & 22.24 & 11.49 \\       
        % \midrule 
        
        %  \bottomrule
        \cline{1-4} \cline{6-9} 
        \end{tabular}
        }
	\vspace{-13pt}

\end{table}

\subsection{Hyper-parameter Studies:} 
\label{sec: hyper}
In this section, we study the effect and selection of the hyper-parameters in IR-FSOD, including the threshold $\tau$ in the semi-supervised RPN, as well as the balance weight $\alpha$ and the scaling factor $\lambda$ in the distance-classifier. For each hyper-parameter, we first select a candidate set by observation and then evaluate their performance on the COCO dataset under the 10-shot setting and one-time FSOD evaluation protocol. The performances at different values of them in Table \ref{tab: param}.
%%
% All the performance is not sensitive when these parameters are in an appropriate range.
%%
The specific analysis is as follows:

\noindent\textbf{Threshold $\tau$}:
Semi-supervised learning on image classification usually requires high thresholds to reduce the incorrect pseudo labels. However, in the two-stage detector, we expect RPN to capture all the potential objects as possible and then eliminate the incorrect proposals by the box classifier. Therefore, the model performs better when $\tau$ is lower and reaches the optimal performance at $\tau = 0.25$.

\noindent\textbf{Balance weight $\alpha$}:
Compared with $\alpha=0.0$ and $\alpha=1.0$, the performances with other values are improved significantly, indicating that the global contrast and the pixel-wise contrast are both valuable.
It reaches the optimal integration performance at $\alpha = \frac{1}{2}$.

\noindent\textbf{Scaling factor $\lambda$}:
In the IR-FSOD, the performance is not affected by the scaling factor since it doesn't affect the confidence ranking of the predicted box. Thus we empirically choose $\lambda=20$ to adjust the sharpness of the prediction distribution.

% \subsection{Visual Result}

\section{Conclusion}
In the few-shot object detection field, the existing methods tend to transfer their model to the detection task by leveraging a fine-tuning process, resulting in many application drawbacks. To tackle the problem, this paper studies in-depth how to get rid of fine-tuning while maintaining the FSOD performance. Through careful study of each module in a general object detector (i.e., Faster R-CNN), this paper builds an instant response few-shot object detector (IR-FSOD) that can accurately detect the object of novel categories while getting rid of the fine-tuning process. To more solidly validate the proposed analysis, we deliberately avoid introducing excessive extra-complexity when designing the improved components. Despite its simplicity, IR-FSOD can reach state-of-the-art performance in both efficiency, precision, and recall. It is noteworthy that our works are built on only Faster R-CNN without other prior methods, so all the approaches are easily compatible with the existing FSOD methods. We hope our studies can inspire future works to explore more powerful few-shot object detectors.

\end{document}